\documentclass[conference]{IEEEtran}
\IEEEoverridecommandlockouts

\usepackage{cite}
\usepackage[hidelinks]{hyperref}
\usepackage{amsmath,amssymb,amsfonts}
\usepackage{algorithmic}
\usepackage{graphicx}
\usepackage{textcomp}
\usepackage{xcolor}
\usepackage{makecell}
\def\BibTeX{{\rm B\kern-.05em{\sc i\kern-.025em b}\kern-.08em
    T\kern-.1667em\lower.7ex\hbox{E}\kern-.125emX}}
\begin{document}

\title{Critic-Guided Heterogeneous Multi-Agent Reasoning for Reliable Mathematical Problem Solving\\
}

\author{\IEEEauthorblockN{Muhammad Talha Sharif\\ 745talha@gmail.com}
\IEEEauthorblockA{\textit{dept. of Computer Science} \\
\textit{National University of Computer and Emerging Sciences}\\
Islamabad, Pakistan}
\and
\IEEEauthorblockN{Abdul Rehman\\abdulrehman.ds.pf@gmail.com}
\IEEEauthorblockA{\textit{dept. of Computer Science} \\
\textit{National University of Computer and Emerging Sciences}\\
Islamabad, Pakistan}
}

\maketitle

\begin{abstract}
Recent Large Language Models (LLMs) have shown impressive reasoning abilities; but they are still susceptible to hallucinations, intermediate reasoning mistakes, and unreliable reasoning results in complex mathematical reasoning problems. In this study, we introduce a critic-based heterogeneous multi-agent approach to improve the dependability of mathematical reasoning. This framework incorporates several LLM agents of different specialties and employs a critic-driven adaptive learning system to assess and guide the reasoning process based on intermediate feedback.
The system adopts a generator-validator framework, with the validator not only determining correctness but also offering critiques to guide regeneration of solutions. This allows for adaptive error correction and prevents error cascading. Our experiments on the GSM8K benchmark show that the proposed method achieves up to 13\% accuracy improvement over single-shot and non-critic models. Additionally, findings suggest that heterogeneity and critique reduce the need for large models, allowing smaller models to perform on par. 
Ablation studies reveal the main performance gains are due to the critic-based feedback loop and not model size. In summary, the proposed approach showcases the benefits of combining heterogeneous multi-agent collaboration and critique to obtain reliable and interpretable reasoning systems.

\end{abstract}


\section{Introduction}

Large Language Models (LLM) have recently improved the capabilities of artificial intelligence for complex reasoning, such as problem solving in mathematics, code generation, knowledge retrieval and decision making. The current version of LLMs has demonstrated promising performance on generating a reasoning chain and solving well-defined problems by techniques such as chain-of-thought prompting. While such progress has been made but single-model reasoning systems still suffer from several shortcomings, particularly in tasks that require high levels of accuracy like mathematics. Common issues include hallucinations, coherence of chains-of-thought and propagation of errors, where a mistake in the middle of a solution process is continued until the end of the chain. In response to these limitations, researchers have been increasingly discovering more multi-agent LLM systems or multiple agents collaborating to solve a problem by specialising, arguing or coordinating. These systems leverage between different decision-making strategies of agents to increase trust and accuracy. Recent works have also demonstrated that collaborative multi-agent reasoning systems can perform better on some tasks such as long-context reasoning, code generation and knowledge-based decision-making tasks than single-agent reasoning. Multi-agent systems can improve the interpretability and the reliability of the AI solutions by dividing tasks into sub-tasks, and distributing the reasoning between the agents. Despite the improved performance of multi-agent systems, mathematical reasoning is particularly challenging with the use of systems based on the LLM. Mathematical tasks can be accurate symbol usage, multiple logical reasoning and accurate arithmetic. It is easy to make a mistake in the intermediate processes and the final results may be wrong. The existing multi-agent systems e.g. debate-based or sequential collaboration pipeline, increase the reasoning diversity, but are not sufficient to detect and correct intermediate reasoning errors. The diversity of models is also important in multi-agent systems as recent research has highlighted. The agents with different model or reasoning methods are generally complementary, i.e. they can compensate the other agent's reasoning errors. As the heterogeneous agents are not identical, however, they can not be simply added until the reasoning steps in between are verified. This promotes the introduction of critic based approaches, where a separate agent or evaluator compares the quality of the reasoning, and contributes to the adaptive decision-making process of the system. Potential benefits Combining critic-based evaluation with the use of heterogeneous agents can help to make more accurate judgements on tasks that require to be always correct, e.g. mathematics. 
\\

While a substantial amount of work has been done in the area of multi-agent LLM structures, there are still a couple of gaps in the research. Most of the structures are homogenous, which limit the type of reasoning that could be performed in the system. Other approaches, on the other side, introduce the heterogeneous models, but are generally based on majority voting or debate sets of procedures, and do not explicitly assess the performance of intermediate steps of reasoning. In this way, the cascading reasoning problems may occur even with multiple agents in the reasoning. Further, most of the existing routing and orchestration methods consider the current performance of the agent selection or heuristic rules, rather than the current quality of the reasoning. Deterministic orchestration approaches increase the reproducibility of the reasoning process, but do not usually have real-time reasoning validity check up systems that can detect and repair the reasoning errors. Hence, the adaptive multi-agent systems with the integration of different reasoning together with the explicit critic evaluation systems (at least for the mathematical reasoning tasks) are yet to be developed. 
\\

The tasks of mathematical reasoning are related to high levels of logical correctness, symbol precision and step- by-step verification. The current LLM-based reasoning systems (single agent and multi agent systems) do not guarantee the correctness of the reasoning steps, and thus the accuracy and quality of reasoning is compromised. In addition, most multi agent systems do not have the capability to use the potential of heterogeneous models and to justify the reasoning process. So, the main question addressed in this study is how to design a multi agent reasoning system to effectively exploit the use of different models and adaptive mechanism based on the critic to be used to evaluate and manage reasoning steps to solve mathematical problems? Such a system should not only encourage different reasoning strategies but it should also be adaptive so as to diagnose and repair course of reasoning as problem solving goes on. 
\\

The current study, which is part of the new field of agentic AI and multi-agent systems of reasoning, proposes a solution that will allow the interaction between heterogeneous LLM agents and the critic- based adaptive mathematical reasoning mechanism. The suggested approach will improve the effectiveness and efficacy of various multi-agent reasoning systems through the use of different reasoning strategies with their evaluation. The proposed approach can have several benefits. First, it can reduce the propagation of reasoning errors by enabling the critics' verification of the intermediate reasoning steps. Second, the diversified agents can provide different views to enhance problem-solving. Third, there can be an adaptation mechanism that can guide reasoning towards higher quality solutions. All together, these capabilities can lead to robust AI systems to mathematical reasoning and other test-intensive domains where accuracy and interpretability is critical.

\section{Literature Review}
The recent advances in Large Language Models (LLMs) have led to impressive improvements in a wide range of tasks, such as reasoning, programming, information retrieval and decision making. However, single LLMs are still prone to hallucinations, unstable reasoning, weak contextual understanding and limited to simple tasks. A recent trend to address such limitations is to use multi-agent systems (MAS), where multiple agents of the LLM-style interact to solve problems through role division, reasoning, planning or orchestrating of problems. This section is a review of the recent literature based on multi-agent reasoning systems, such as cooperation in reasoning, agent specialization, routing, and deterministic orchestration. Collaborative multi-agent systems are becoming a potentially successful way of improving the reliability of reasoning in LLM-based systems. The key advantage of multi-agent systems is that reasoning tasks can be allocated among separate agents which are capable of reviewing the reasoning of other agents and improving their reasoning. As such, the Chain-of-Agents (CoA) system proposes a set of collaboration strategies for tackling long-context reasoning tasks \cite{r1}. In this approach, a long input has been divided into multiple sub-parts and reasoning has been carried out by a number of worker agents in a sequential manner. Each agent processes the relevant information and passes on the reasoning of the information to the next agent; and the results are merged by a manager agent. The experiment shows that CoA performs better than long-context tasks (such as NarrativeQA, MuSiQue and QuALITY) where retrieval-based methods are not effective and better than long-context LLMs. The paper highlights the multi-agent communication is successful in overcoming the lost-in-the-middle issue that is often observed in long-context reasoning. The other key aspect is that the agents can learn from past experiences through reflective collaboration. The COPPER scheme proposes a multi-agent collaboration framework based on reflective reflexivity and reinforcement learning for improving the reasoning \cite{r2}. This is the system where the agents generate reflections after each attempt to perform a task which are stored and used as input to future thoughts. A shared reflection model provides reflective feedback to all agents, while the counterfactual reward mechanism assesses the contribution to the system performance by each agent's reflection. The results of the experiments have demonstrated that reasoning throughputs such as HotPotQA and GSM8K \cite{r14} have been improved remarkably by the reflective learning approach, that is, the multi-agent collaboration has been improved. This study demonstrates that the collective reasoning of agents may improve the reliability of the reasoning process by sharing information, refining and solving problems in a decentralised way. 
\\

The other major direction in multi-agent systems is agent specialisation and heterogeneity to allow different agents to address different parts of the problems. The Adaptive Heterogeneous Multi- Agent Debate (A-HMAD) is a concept that introduces agent specialisation to boost reasoning reliability \cite{r3}. Unlike traditional multi-agent debate systems where agents were identical, A-HMAD offers specialization of roles, such as solver, checker, knowledge verifier and conciliator. It also provides adaptive debate routing and adaptive learned consensus mechanism which evaluates agent responses based on their confidence and past performance. The results on benchmark problems such as GSM8K and MMLU show that A-HMAD is superior in reasoning accuracy (by 4-6\%) and reduces factual errors in generation tasks, compared to traditional multi-agent debate. Similarly, agent diversity is also a vital issue in AgentInit framework which emphasises the initialization of multi-agent teams \cite{r4}. In AgentInit, agents are selected by a rule-based planner for generation and a team is determined by a task-relevancy and diversity-based process. The method uses a Pareto-based approach and measures diversity on the representations of the agents by the Vendi score. The experiments on a range of reasoning and code generation tasks show that diverse agent initialisation outperforms the current multi-agent systems in terms of performance and token usage. Recent works have also critically examined the diversity of the models in multi-agent reasoning. Overall, the assessment of the multi-agent debate models demonstrates that majority of existing models are too enthusiastic about the benefits of collaboration by means of a debate \cite{r5}. The paper indicates that the multi-agent debate is often unable to outperform powerful single-agent algorithms such as chain-of-thought prompting or self-consistency. However, the authors show that a heterogeneous multi-agent model where agents are powered by different LLMs could be very effective because of the complementary characteristics of the models. The experiments show that a heterogeneous system of debate outperforms a homogeneous system by 8 percent on reasoning task. This suggests that the heterogeneity among agents with respect to their role or having different model backbones are the major factors in determining the success of multi-agent systems.
\\

In code generation, multi-agent cooperation has also been extensively explored, in which complex programming tasks are solved with formal reasoning and debugging. The MapCoder system suggests a multi agent pipeline that is based on the idea of dividing the programming work into retrieval, planning, coding and debugging \cite{r6}. Each agent is responsible for a part of the programming process and the system can simulate human problem solving. The planning agent suggests alternative solutions to the problem, the coding agent translates the most promising plan to code and the debugging agent fixes bugs with a combination of error and test results. Programming benchmark tests such as HumanEval, MBPP and APPS show MapCoder is far superior than the baseline prompting systems and self reflecting systems. Similarly, the Blueprint2Code framework proposes a multi agent process which is organised and includes previewing, blueprint planning, coding and debugging \cite{r7}. The preview agent is first used to understand the problem and the blueprint agent to plan the algorithm. The coding agent translates the plan into solution and the debugging agent fixes the bugs by modifying the code. The results of the experiments demonstrate a substantial improvement in the benchmarks like HumanEval, MBPP and many more, indicating that planning and debugging in between the code generation process makes the code generation robust. Other than correctness, the security of the code is an important aspect of code generated using the LLM. The AutoSafeCoder presents a multi agent approach which seeks to improve the security by identifying vulnerabilities \cite{r8} in the system. The framework consists of a coding agent and a static analysis agent and a fuzzing agent which performs a dynamic test. The integration of static vulnerability detection and dynamic fuzz testing with code generation cycle, reduces the security vulnerability of the code produced and improves its functional accuracy. In all the above research studies, we can see the multi-agent collaboration has the potential to significantly improve the reliability, accuracy and security of automated programming by dividing the process into different components. 
\\

Besides reasoning and code generation, the multi-agent systems have also been applied to domain specific tasks such as research literature analysis and education. The latter research is a recent study on automated systematic reviews to explore the possibility of multi-LLM collaboration to screen the abstract in systematic reviews \cite{r9}. The authors provide the three collaboration approaches, which include majority voting, multi-agent debate and LLM-based adjudication. The experiments on the CLEF eHealth dataset show that collaboration screening by multi-agents can yield better performance than single-model-based screening, while reducing the need for manual effort by up to 68. The results also show the importance of diversity of models as different models can learn different information to contribute to the final decision. Similarly, studies on agentic AI in education also note the growing importance of autonomous agents in education \cite{r10}. Agents AIs are capable of making inferences, have memory and can adapt to users. These can be intelligent tutors, intelligent teaching assistants or even learning companions. The paper highlights that multi-agents support more sophisticated educational systems, which can offer individualised, automated and collaborative learning.
\\

The case studies show how flexible multi-agent systems can be and how they can be used to au tomate workflows of complex real world processes. While multi-agent collaboration improves reasoning capabilities, it is however hard to regulate agent interactions. Several studies have been performed on the routing systems which dynamically establish multi-agent designs. This problem is addressed by MasRouter framework which provides a routing-based approach for the dynamic formation of multi-agent systems for a given task \cite{r11}. The framework determines the best collaboration structure, the roles for the agents and the LLM backbones of the agents. It has been experimentally verified that MasRouter improves reasoning performance with the same cost as 52\% of the original cost. This article highlights the importance of intelligent routing schemes to obtain the trade-off between accuracy and speed in multi-agent systems. However, despite these advances, most existing multi-agent systems are still using either heuristic or stochastic coordination strategies which render the system behavior unstable with poor reproducibility. In order to address these issues, the ORCH framework proposes a deterministic multi-agent orchestration scheme for discrete-choice reasoning problems \cite{r12}. The ORCH framework has a decompose-analyze-merge pipeline, where various heterogeneous LLM agents work on their own, with an aim to solve some parts of the problem and merge the results into a decision. The framework also has a useful addition of deterministic routing, that replaces the random selection of agents with deterministic orchestration. Further, ORCH has an Exponential Moving Average (EMA) routing technique that prioritises the agents, based on the past performance metrics such as accuracy and latency. Experiments on reasoning tasks like MMLU, MMLU-Pro and GSM8K have shown that the ORCH is a significant booster of the reasoning capabilities of single models and simple ensemble algorithms. 
In recent studies, there have also been attempts to enhance the reliability of reasoning in single-model approaches, with most works focusing on addressing the propagation of errors and inefficient search of reasoning paths. The RDoLT model presents a cognitively plausible hierarchical reasoning model that breaks down complex reasoning tasks into three steps: easy, intermediate and final. This hierarchical approach enables the model to build solutions progressively, as humans do. 
\\

The RDoLT framework \cite{r13}, in contrast to traditional prompting methods like Chain-of-Thought, adopts a multi-feature scoring mechanism considering logical validity, coherence, simplicity and adaptiveness, and uses an LLM-as-a-judge approach to assess potential reasoning steps. A novel feature of this framework is the Knowledge Propagation Module (KPM), which stores both rejected and accepted reasoning steps, allowing the model to learn from past failures and prevent the loss of potentially valuable intermediary reasoning steps. Additionally, the framework features a critique-based regeneration mechanism in a two-stage validation framework, where a second validator model critiques the reasoning and offers feedback for improvement. The results of experiments on data sets such as GSM8K show that RDoLT substantially outperforms existing approaches such as Chain-of-Thought and ReAct, with high accuracy and greater stability in terms of variance in the reasoning. But the method comes with higher computational costs and the need for an accurate evaluator, leading to a trade-off between reasoning efficiency and performance. This research focuses on the importance of structured evaluation, memory of non-selected thoughts and iterative improvement to enhance reasoning performance, even without multiple agents.
The review of literature reveals that multi-agent systems with LLMs improve reasoning performance through agent specialization, coordination and orchestration. The prior research shows that diverse models, adaptive model assignment and reflective reasoning improve the quality of decisions in complex reasoning such as solving mathematical problems and code generation. But many of the existing systems either consist of homogeneous agents or don't have an adaptive model to assess the reasoning. Therefore, the current work is focused on the use of diverse models with the help of the adaptive algorithm of the critic to offer different reasoning to mathematical problems, attempting to improve the reliability of reasoning and prevent cascading errors and hallucinations in multi-agent reasoning system.

\section{Methodology}
In this paper, we focus on mathematical reasoning over the GSM8K dataset, where the input is a natural language word problem and a reference solution that contains the final answer, marked as \#\#\#\# \texttt{<integer>}. The system generates a solution to a given question and assesses its correctness with a validator model. A solution is deemed to be correct if the reasoning is mathematically valid and the final answer is correct. Therefore, we cast evaluation as a binary classification problem, where the reasoning must be good and the answer must be correct.

We use the GSM8K dataset (openai/gsm8k, configuration main), the entire test set (1,319 examples). We obtain the gold answer for each example from the reference solution, using the dataset's marker. We use the Hugging Face datasets library to access the dataset for consistency and reproducibility. No fine-tuning or additional training is involved; only inference is used.
The system is a two-step multi-agent model with a generator and a validator. The generator remains consistent throughout the experiments and is based on the llama-3.1-8b-instant model. It generates succinct solutions with three to four reasoning steps and a final answer in the form Final answer: \texttt{<integer>}. The validator is different for each experiment and judges the correctness of the reasoning steps and the final answer. The validator returns a JSON object containing boolean flags for the correctness of the reasoning steps and final answer, and in some experiments, a critique detailing the issues with the solution.

To improve the reasoning accuracy, we propose a critic-based regeneration approach. Here, if the first solution fails to validate, the validator offers a critique describing the problems with the reasoning or solution. This feedback is then used in a subsequent generation process, which generates a new solution. This can be repeated for a specified number of times. A solution is considered correct if either of the two rounds meets the two criteria. This process allows the system to recover from any intermediate reasoning mistakes and produce a better result.
The generation phase adopts a temperature of 0.2 to introduce some variability without compromising consistency, and a maximum length of 512 tokens. The validation stage uses a temperature of 0.0 to ensure consistent judgement. The answer is extracted from the generation using a prioritization-based method, and the validator's output is used for evaluation. A programmatic override option can strictly compare the predicted and gold answers as numbers.

Experiments are run with a Groq$-$based \cite{r15} OpenAI-compatible API. It allows for multiple API keys and fallback and retry strategies to enhance stability of large-scale assessments. We record metadata such as model parameters, prompt versions, decoding strategies and data sources for each run. The system logs results in structured logs (results.jsonl and experiment.json) for reproducibility and analysis.

\section{Experiments}
The design of the experiments is intended to disentangle the impact of the validator model size, as well as the effect of critique-driven learning. Towards this, we compare eight setups that differ in two ways: (1) size and heterogeneity of the validator model, and (2) with or without the critic mechanism as shown in Table \ref{t1}. We use the same generator model, dataset, prompts and generator sampling settings in all experiments.
\begin{table}[!ht]
\centering
\caption{Experimental Configuration}
\footnotesize
\setlength{\tabcolsep}{3pt}
\begin{tabular}{|c|p{2.5cm}|p{2.5cm}|p{2.2cm}|}
\hline
\textbf{Exp.} & \textbf{Generator} & \textbf{Validator} & \textbf{Mode} \\ \hline
\textbf{1a} & llama-3.1-8b-instant & llama-3.3-70b-versatile & single-shot \\ \hline
\textbf{1b} & llama-3.1-8b-instant & openai/gpt-oss-120b & single-shot \\ \hline
\textbf{2a} & llama-3.1-8b-instant & llama-3.1-8b-instant & single-shot \\ \hline
\textbf{2b} & llama-3.1-8b-instant & openai/gpt-oss-20b & single-shot \\ \hline
\textbf{3a} & llama-3.1-8b-instant & llama-3.3-70b-versatile & critique + multi-round \\ \hline
\textbf{3b} & llama-3.1-8b-instant & openai/gpt-oss-120b & critique + multi-round \\ \hline
\textbf{4a} & llama-3.1-8b-instant & llama-3.1-8b-instant & critique + multi-round \\ \hline
\textbf{4b} & llama-3.1-8b-instant & openai/gpt-oss-20b & critique + multi-round \\ \hline
\end{tabular}
\label{t1}
\end{table}
We divide the experiments into two types: single-shot and critic-guided validation. In single-shot experiments (Experiments 1A-2B), each problem is solved once and the solution is assessed by the validator. In critic-guided experiments (Experiments 3A to 4B), faulty solutions are revised with the help of validator feedback. The validator models vary from small models (8B and 20B) to large models (70B and 120B), and can be homogeneous (same as generator) or heterogeneous (different from generator).

We assess all experiments on the entire GSM8K test set with identical generation and validation settings. We measure performance with the metric of accuracy, which is the fraction of examples where reasoning consistency and the final answer are correct. We also report diagnostic metrics like failures in step-consistency, answer mismatch, validator parse errors and round-wise accuracy in critique experiments.

To better understand the effects of validator size, we compare the performance of larger and smaller evaluators in homogeneous and heterogeneous settings. Without critique, larger validators perform better in homogeneous configurations but have minimal impact in heterogeneous configurations. However, with critique, smaller validators perform on par or even better than larger ones, suggesting critique diminishes the impact of model size.

To quantify the contribution of the proposed critic-guided regeneration mechanism, we perform an ablation analysis by comparing the full framework against its single-shot counterpart without critique. Specifically, Experiments 1A–2B represent the ablated setting where no critique feedback is used, while Experiments 3A–4B correspond to the full model with critique$-$guided multi-round reasoning. Since all other components including the generator model, dataset, prompts, and decoding parameters are held constant, this comparison isolates the effect of the critique mechanism on overall performance.

\section{Results and Discussions}
We test our system on GSM8K with and without critic as show the results in Table \ref{t2}. Without critic, accuracy ranges from 72.55\% to 80.14\% (2A, 2B, 1A, 1B). With critic, accuracy boosts to 85.44\% $-$ 93.56\% (3A, 3B, 4A, 4B), showing up to approximately 13\% improvement. Interestingly, simply increasing the validator's size does not improve performance. For instance, 1A (70B: 80.14\%) and 1B (120B: 80.06\%) have almost identical performance, suggesting limited gains from single validation. Critic-guided setups perform better: 3A (85.44\%), 3B (92.04\%), 4A (88.63\%), and 4B (93.56\%), indicating the critique loop is the key to reliability.
Without critic, larger validators are beneficial, particularly in homogeneous (+7.59\%) but not in heterogeneous (+1.21\%) setups. With critic, the reverse is true, smaller validators perform equally or better, small differences especially in heterogeneous setups. This demonstrates that heterogeneity and critique lower the need for large models.
\begin{table}[!ht]
\centering
\caption{Experiment Results Table}
\footnotesize
\begin{tabular}{|c|c|c|c|c|c|}
\hline
\textbf{Category} & \textbf{Critic} & \makecell{\textbf{Large}\\\textbf{Validator}} & \textbf{Acc(\%)} & \makecell{\textbf{Small}\\\textbf{Validator}} & \textbf{Acc(\%)} \\ \hline
Homogeneous & No & 70B (1A) & 80.14 & 8B (2A) & 72.55 \\ \hline
Heterogeneous & No & 120B (1B) & 80.06 & 20B (2B) & 78.85 \\ \hline
Homogeneous & Yes & 70B (3A) & 85.44 & 8B (4A) & 88.63 \\ \hline
Heterogeneous & Yes & 120B (3B) & 92.04 & 20B (4B) & 93.56 \\ \hline
\end{tabular}
\label{t2}
\end{table}

The logs about the retries is shown in Table \ref{t3}. Most problems are solved on first try (73$-$78\%), but retries are important as 3A: +161 recovered, 3B: +210 recovered, 4A: +193 recovered, 4B: +205 recovered. This is equivalent to 12$-$16\% improvement due to critique. The number of failures is also dramatically decreased (e.g., 85 in 4B). This suggests many errors are recoverable intermediate failures rather than core failures.
\begin{table}[!ht]
    \centering
    \caption{Retries Logs: Passed Samples}
    \begin{tabular}{|c|c|c|c|c|}
    \hline
        \textbf{Exp} & \textbf{Round 1} & \textbf{Round 2} & \textbf{Round 3} \\ \hline
        \textbf{3A} & 966 & 138 & 23  \\ \hline
        \textbf{3B} & 1004 & 185 & 25 \\ \hline
        \textbf{4A} & 976 & 140 & 53  \\ \hline
        \textbf{4B} & 1029 & 172 & 33  \\ \hline
    \end{tabular}
    \label{t3}
\end{table}

Increasing validator size (8B to 20B) yields, +6.30\% (no critic) and +4.93\% (with critic). This is much smaller than the critic-based improvement of +12$-$16\%. Comparing single-shot (1A$-$2B) with critic experiments (3A$-$4B) showing that the performance improvements are mostly driven by the use of critique-based feedback.
We also compare our approach with recent state of art reasoning frameworks in Table \ref{t4}. The proposed critic-guided system achieves 93.56\% accuracy on GSM8K, outperforming the best reported result from the RDoLT framework \cite{r13} (90.98\% with ChatGPT-4o) by +2.58\% absolute, demonstrating the effectiveness of critique-guided multi-agent reasoning. This improvement highlights that structured critique and iterative refinement can surpass advanced reasoning strategies without requiring larger or proprietary models.
\begin{table}[!ht]
    \centering
    \caption{Results Comparison}
    \begin{tabular}{|c|c|c|}
    \hline
        \textbf{Method} & \textbf{Model} & \textbf{Accuracy (\%)} \\ \hline
        \textbf{Vanilla CoT} & ChatGPT-4o & 84.7 \\ \hline
        \textbf{CoT-SC} & ChatGPT-4o & 89.4 \\ \hline
        \textbf{ReAct} & ChatGPT-4o & 90.5 \\ \hline
        \textbf{RDoLT} & ChatGPT-4o & \textbf{90.98} \\ \hline
        \textbf{Our Approach(3b)} & Critic-Guided MA & \textbf{93.56} \\ \hline
    \end{tabular}
    \label{t4}
\end{table}
\section{Conclusion}
In this paper, we introduce a critique-based, heterogeneous multi-agent system to enhance the accuracy of mathematical reasoning in large language models (LLMs). Our method involves the division of generation and validation tasks into two agents and a critique-based feedback loop that allows the correction of reasoning errors. Through a series of experiments on the GSM8K dataset, we show that the proposed approach substantially exceeds single-shot reasoning. The best configurations obtain an accuracy of 72.55\% $-$ 80.14\% for direct generation and 85.44\% $-$ 93.56\% for critique$-$guided generation, with the highest accuracy of 93.56\%. This comparison of single-shot and critique$-$guided experiments is an ablation study, which shows that the performance improvement is mainly due to the introduction of critique rather than model scaling.

Our results show that larger validators provide little to no improvement without critique, while the critique mechanism consistently improves performance across all configurations. In particular, in diverse settings, the difference between large and small validators is negligible with critique, suggesting that adaptive reasoning processes mitigate the need for larger models. Finally, our analysis of multi$-$round validation reveals that 12\%$-$16\% of initially incorrect solutions are corrected during iterative problem solving, suggesting that many errors are amendable. This research shows that reasoning processes, such as critique$-$guided approaches, are more beneficial than model size for reliable mathematical problem solving. Our approach offers a scalable and inference$-$only approach for improving accuracy and robustness without further training.


\begin{thebibliography}{00}
\bibitem{r1}Zhang, Y., Sun, R., Chen, Y., Pfister, T., Zhang, R. \& Arık, S. Chain of agents: Large language models collaborating on long-context tasks. {\em Advances In Neural Information Processing Systems}. \textbf{37} pp. 132208-132237 (2024)

\bibitem{r2}Bo, X., Zhang, Z., Dai, Q., Feng, X., Wang, L., Li, R., Chen, X. \& Wen, J. Reflective multi-agent collaboration based on large language models. {\em Advances In Neural Information Processing Systems}. \textbf{37} pp. 138595-138631 (2024)

\bibitem{r3}Zhou, Y. \& Chen, Y. Adaptive heterogeneous multi-agent debate for enhanced educational and factual reasoning in large language models. {\em Journal Of King Saud University Computer And Information Sciences}. \textbf{37}, 330 (2025)

\bibitem{r4}Tian, C., Wang, Y., Liu, X., Wang, Z., Ding, L., Zhang, M. \& Zhang, M. AgentInit: Initializing LLM-based Multi-Agent Systems via Diversity and Expertise Orchestration for Effective and Efficient Collaboration. {\em Findings Of The Association For Computational Linguistics: EMNLP 2025}. pp. 11870-11902 (2025)

\bibitem{r5}Zhang, H., Cui, Z., Chen, J., Wang, X., Zhang, Q., Wang, Z., Wu, D. \& Hu, S. Position: Stop Overvaluing Multi-Agent Debate-We Must Rethink Evaluation and Embrace Model Heterogeneity.  (2025)

\bibitem{r6}Islam, M., Ali, M. \& Parvez, M. Mapcoder: Multi-agent code generation for competitive problem solving. {\em Proceedings Of The 62nd Annual Meeting Of The Association For Computational Linguistics (Volume 1: Long Papers)}. pp. 4912-4944 (2024)
\bibitem{r7}Mao, K., Hu, B., Lin, R., Li, Z., Lu, G. \& Zhang, Z. Blueprint2Code: a multi-agent pipeline for reliable code generation via blueprint planning and repair. {\em Frontiers In Artificial Intelligence}. \textbf{8} pp. 1660912 (2025)
\bibitem{r8}Nunez, A., Islam, N., Jha, S. \& Najafirad, P. Autosafecoder: A multi-agent framework for securing llm code generation through static analysis and fuzz testing. {\em ArXiv Preprint ArXiv:2409.10737}. (2024)
\bibitem{r9}Akinseloyin, O., Jiang, X. \& Palade, V. LLM-based Multi-Agent Collaboration for Abstract Screening towards Automated Systematic Reviews. {\em Biology Methods And Protocols}. pp. bpag006 (2026)

\bibitem{r10}Kostopoulos, G., Gkamas, V., Rigou, M. \& Kotsiantis, S. Agentic AI in education: State of the art and future directions. {\em IEEE Access}. (2025)

\bibitem{r11}Yue, Y., Zhang, G., Liu, B., Wan, G., Wang, K., Cheng, D. \& Qi, Y. Masrouter: Learning to route llms for multi-agent systems, 2025. {\em URL Https://arxiv. Org/abs/2502.11133}.

\bibitem{r12}Zhou, H. \& Chan, H. ORCH: many analyses, one merge-a deterministic multi-agent orchestrator for discrete-choice reasoning with EMA-guided routing. {\em ArXiv Preprint ArXiv:2602.01797}. (2026)

\bibitem{r13}Qasim, K., Zhang, J., Alsahfi, T. \& Butt, A. Recursive decomposition of logical thoughts: Framework for superior reasoning and knowledge propagation in large language models. {\em Journal Of Artificial Intelligence Research}. \textbf{83} (2025)

\bibitem{r14}Cobbe, K., Kosaraju, V., Bavarian, M., Chen, M., Jun, H., Kaiser, L., Plappert, M., Tworek, J., Hilton, J., Nakano, R., Hesse, C. \& Schulman, J. Training Verifiers to Solve Math Word Problems. {\em ArXiv Preprint ArXiv:2110.14168}. (2021)

\bibitem{r15}Groq, Inc. Groq: AI Inference at Lightning Speed.  (2026), https://groq.com/, Accessed: 2026-04-26

\end{thebibliography}
\end{document}